\documentclass[letterpaper, 10 pt, conference]{ieeeconf}  

\IEEEoverridecommandlockouts                              

\usepackage{graphics} 
\usepackage{epsfig} 
\usepackage{mathptmx} 
\usepackage{times} 
\usepackage{amsmath} 
\usepackage{amssymb}  

\usepackage{comment}
\usepackage{booktabs}
\usepackage{comment}
\usepackage{cite}

\usepackage{color}
\usepackage[dvipsnames]{xcolor}
\usepackage{hyperref}
\let\labelindent\relax
\usepackage{enumitem}
\usepackage{color, colortbl}
	
\definecolor{LightCyan}{rgb}{0.88,1,1}
\definecolor{highlight}{rgb}{0.824,0.976,0.824}
\definecolor{darkgreen}{rgb}{0,0.45,0}
\usepackage{colortbl}

\usepackage{pifont}
\usepackage{newunicodechar}
\newunicodechar{✓}{\ding{51}}
\newunicodechar{✗}{\ding{55}}
\usepackage{wrapfig}

\usepackage[capitalize]{cleveref}
\crefname{section}{Sec.}{Secs.}
\Crefname{section}{Section}{Sections}
\Crefname{table}{Table}{Tables}
\crefname{table}{Tab.}{Tabs.}

\title{\LARGE \bf
Can Reasons Help Improve Pedestrian Intent Estimation? 

A Cross-Modal Approach
}

\author{Vaishnavi Khindkar$^{\star}$, Vineeth Balasubramanian$^{\rceil}$, Chetan Arora$^{\ddag}$ Anbumani Subramanian$^{\dag}$, C.V. Jawahar$^{\dag}$\\
\thanks{$^{\star \rceil \dag \ddag}$~The authors are with the Center for Visual Information Technology (CVIT) Lab, IIIT Hyderabad, IIT Delhi and IIT Hyderabad, India.
        {\tt\scriptsize $^{\star}$ vkhindkar@gmail.com},
        {\tt\scriptsize $^{\rceil}$ vineethnb@cse.iith.ac.in},        
        {\tt\scriptsize $^{\dag}$ \{anbumani,jawahar\}@iiit.ac.in},  {\tt\scriptsize $^{\ddag}$chetan@cse.iitd.ac.in}}%
}

\begin{document}

\maketitle
\thispagestyle{empty}
\pagestyle{empty}


\begin{abstract}
With the increased importance of autonomous navigation systems has come an increasing need to protect the safety of Vulnerable Road Users (VRUs) such as pedestrians. Predicting pedestrian intent is one such challenging task, where prior work predicts the binary cross/no-cross intention with a fusion of visual and motion features. However, there has been no effort so far to hedge such predictions with human-understandable reasons. We address this issue by introducing a novel problem setting of exploring the intuitive reasoning behind a pedestrian's intent. In particular, we show that predicting the `WHY' can be very useful in understanding the `WHAT'. To this end, we propose a novel, reason-enriched PIE++ dataset consisting of multi-label textual explanations/reasons for pedestrian intent. 
We also introduce a novel multi-task learning framework called \textsc{MINDREAD}, which leverages a cross-modal representation learning framework for predicting pedestrian intent as well as the reason behind the intent. Our comprehensive experiments show significant improvement of 5.6\% and 7\% in accuracy and F1-score for the task of intent prediction on the PIE++ dataset using \textsc{MINDREAD}. We also achieved a 4.4\% improvement in accuracy on a commonly used JAAD dataset. Extensive evaluation using quantitative/qualitative metrics and user studies shows the effectiveness of our approach.

\end{abstract}

\begin{figure*}
    \center
    \includegraphics[width=0.87\textwidth]{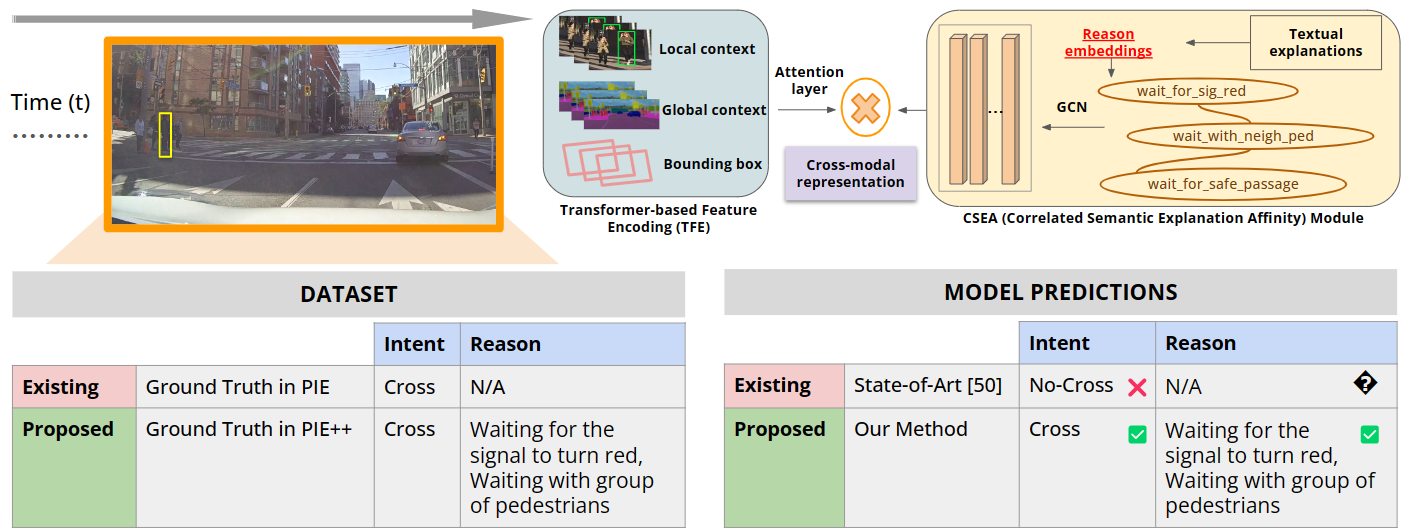}
    \vspace{-1mm}
    \caption{Illustration of our overall objectives. Given a sample scene from the PIE dataset \cite{rasouli2019pie}, we study the usefulness of ``WHY" (reason) for analyzing ``WHAT" (pedestrian intent) by creating a reason-enriched PIE++ dataset. While state-of-the-art \cite{yao2021coupling} makes an incorrect prediction on this scene, \textsc{MINDREAD}, our cross-modal representation learning approach (architecture on top), correctly predicts the crossing intent with the support of a corresponding reason.}
    \label{fig:teaser}
    \vspace{-2mm}
\end{figure*}

\section{Introduction}
\label{sec:intro}
Autonomous navigation and Advanced Driver Assistance Systems (ADAS) have seen significant interest and developments in recent years from the computer vision research community in both industry and academia. The development of newer deep learning architectures and models, as well as the availability of datasets for various tasks such as object detection and semantic segmentation for road scenes \cite{Cordts_2016_CVPR, Yu_2020_CVPR, 6248074}, have catalyzed the development of deployable solutions for these tasks in autonomous navigation systems. However, an important aspect of such systems and vehicles is how they understand and interact with Vulnerable Road Users (VRUs) like pedestrians \cite{reyes2022vulnerable}. There have been extensive studies in recent years by government and custodial organizations to protect the safety of VRUs, especially pedestrians, as we develop more advanced autonomous navigation systems \cite{vrulink1,vrulink3}. In this work, we focus on this key aspect of such systems -- understanding pedestrian behavior in road scenes.

Prior efforts in recent years \cite{rasouli2019pie, liu2020spatiotemporal, lorenzo2021intformer, piccoli2020fussi, neogi2020context, saleh2019real, 8317941} have attempted the problem of predicting pedestrian intent using motion models, poses and trajectories. While each of these efforts have been vital to progress in this field, an important aspect in understanding pedestrian behavior that has not been considered in these methods is -- reasoning. Studies on human behavior have shown reasoning to be an important link between user belief, intention and behavior \cite{westaby2005behavioralreasoning}. We specifically seek to address this perspective in this work by asking: ``Can reasoning help improve pedestrian intent prediction?''
We answer the WHAT, and study the importance of WHY in predicting the WHAT. We illustrate this with a sample result of our method in Fig \ref{fig:teaser}. As shown in the figure, the pedestrian has an intention to cross the road (also called `\textit{crossing}' intention in this work and related literature). This is supported by the reason that `\textit{the pedestrian is waiting for the signal to turn red, and is waiting with a group of pedestrians}'. Current state-of-the-art \cite{yao2021coupling} incorrectly predicts the intent as `\textit{no-crossing}' (no intention to cross), while our method -- which considers the reason in its prediction -- predicts the correct pedestrian intent. Providing a window into such reasoning can also be helpful to a driver in increasing the trust in the model that predicts pedestrian intent. We use the terms reasons and explanations interchangeably at times in this work, although explanations refer to the interpretation of pedestrian intent in our context and not model interpretability. 

\vspace{-3pt}
Motivated by such a reasoning-driven understanding of pedestrian intent, we make two significant contributions in this work: (i) We study the usefulness of reason in predicting pedestrian intent on road scenes, and develop a new methodology based on cross-modal representation learning and attention that uses vision and language modules towards the stated objective; and (ii) We obtain enriched annotations on reasons for pedestrian intent on the benchmark PIE dataset \cite{rasouli2019pie}, and thus develop the first pedestrian reason+intent prediction dataset which we call PIE++ (we describe in detail why we choose PIE as the dataset for enrichment with reason annotations in Sec \ref{pie++}). 
PIE++ consists of human-referenced, multi-label explanation annotations for all the $1842$ pedestrians, and could potentially serve as a useful resource for researchers in the community. 
Based on a meticulous study (discussed in Sec \ref{pie++}), we include commonly understood reasons for pedestrian intent while crossing or not crossing the streets in PIE++, such as: waiting for safe passage, pedestrians doing their work, and so on. These reasons, beyond explaining pedestrian intent, may also provide a driver with a better understanding of the road scene. 

\vspace{-3pt}
From a method perspective, we hypothesize that cross-modal representation learning that includes vision and language modules helps capture pedestrian intent better, compared to prior works that use only visual features. In particular, textual explanations (annotated reasons in PIE++) that are rich in semantic descriptions provide more information than binary labels of crossing intention of a pedestrian, and thus help support better intent prediction. We hence introduce a multi-task formulation based on cross-modal representation learning to predict both pedestrian intent as well as the reason behind the intent. 
We propose a novel semantic correlation module to capture the relationships between the reason text embeddings, and fuse this with the visual spatio-temporal features through an attention-based mechanism to obtain cross-modal representations.
Our overall method, which we call \textsc{MINDREAD} (cross-Modal representatIon learNing moDel for REAsoning peDestrian-intent), consists of three modules: (i) Correlated Semantic Explanation Affinity (CSEA); (ii) Transformer-based Feature Encoding (TFE); and (iii) Attention-based cross-modal representation learner. CSEA assesses the co-existence of explanations to obtain an affinity matrix, and then uses a Graph Convolutional Network to obtain embeddings for textual explanations. These are fused with spatiotemporal features obtained from local visual context, global visual context and bounding boxes in the TFE module using an attention mechanism. 
The textual-explanation affinity modeling is a critical component of our framework as it helps leverage the reason for the pedestrian intent, thereby improving intent prediction. 
Following are the key contributions of our work:
\vspace{-2mm}
\begin{itemize}[leftmargin=*]
\setlength\itemsep{-0.25em}
\item We present a novel perspective of considering reason as a means to predict pedestrian intent and a multi-task formulation to simultaneously predict pedestrian intent and its corresponding reason. 
\item We provide PIE++, enrichment of a benchmark dataset for intuitively reasoning pedestrian intent, comprising human-referenced multi-label explanation/reason annotations rich in semantic descriptions, providing more information than the binary crossing intention labels alone. 
\item We propose a new cross-modal representation learning framework, \textsc{MINDREAD}, 
that exploits semantic correlations between textual embeddings and attention-based fusion with spatio-temporal features for intent prediction. 
\item We perform a comprehensive suite of experiments to study our method and dataset. Our results show significant improvement of 5.6\% and 7\% in accuracy and F1-score for intent prediction on the PIE++ dataset using \textsc{MINDREAD}. We also achieve 4.4\% improvement over state-of-the-art in accuracy on the JAAD dataset \cite{rasouli2017they}. 
\end{itemize}



\section{Related Work}
\label{sec:related work}

We herein discuss earlier efforts related to pedestrian intent prediction, the focus area of this work. Other related work such as action prediction \cite{Rasouli_2021_ICCV, rasouli2020pedestrian, 9561107, Kotseruba_2021_WACV, yao2021coupling, zhai2022social, rasouli2022multi} have different objectives such as generating future
frames, predicting action types, measuring confidence in
event occurrence, and forecast object motion moreover focusing on future action anticipation. Textual explanations \cite{vedantam2017context, hendricks2018grounding, xu2015show, lecun2015deep, kim2018textual, chen2021psi, Ge_2021_CVPR, Bao_2021_ICCV}  have sometimes been used in earlier work for obtaining insights on a neural network's understanding of images or scenes. Cross-modal representation learning \cite{radford2021learning, Yuan_2021_CVPR, Choi_2021_CVPR} works learn latent
semantic representations using multiple modalities and is
applicable in diverse deep-learning tasks including image
captioning, cross-modal retrieval, visual question-answering classification, and detection. 

\noindent \textbf{Intent vs. Action?}
Existing action prediction methods \cite{Rasouli_2021_ICCV, rasouli2020pedestrian, 9561107, Kotseruba_2021_WACV, yao2021coupling, zhai2022social, rasouli2022multi} have different objectives such as generating future frames, predicting action types, measuring confidence in event occurrence, and forecast object motion. We hence note that pedestrian action prediction and pedestrian intent prediction (which is understanding the event before the action) are two different streams of work as defined in the literature and followed by the community in this space. The results from action prediction works are not directly comparable with intent prediction results, given the differences between the two predefined tasks. We also note that our work does not change any pre-existing definitions of pedestrian intent defined in \cite{rasouli2019pie}. Also, following \cite{rasouli2019pie}, we consider the intent only up to a critical point (i.e. frames preceding the crossing point).

\noindent \textbf{Pedestrian Intent Prediction.}
Prior works \cite{rasouli2019pie, liu2020spatiotemporal, lorenzo2021intformer, piccoli2020fussi, neogi2020context, saleh2019real, 8317941, kim2020pedestrian, sui2021joint} typically model pedestrian intent as motion and use poses or trajectories to predict future goals. PIEint \cite{rasouli2019pie} encodes past visual features, and then concatenates them with bounding boxes to predict intent. A framework based on a pedestrian-centric graph is proposed in \cite{liu2020spatiotemporal} to uncover spatio-temporal relationships in the scene. Other works such as \cite{piccoli2020fussi} study early, late, and combined (early+late) fusion mechanisms to exploit skeletal features and improve intent prediction. A conditional random field-based approach is used in \cite{neogi2020context} for early prediction of pedestrian intent. In the current state-of-the-art for this problem, the recently proposed method in \cite{yao2021coupling} predicts future pedestrian actions and uses predicted action to detect present intent. While each of these efforts have shown promising results, we approach this problem from a different perspective of reasoning. 
In particular, we employ rich semantic descriptions (reasons) to evaluate the potential of obtaining better prediction of pedestrian intent, and thereby also assist an end-user with a human-interpretable reason. Also, unlike the abovementioned works that utilize only vision features, ours is a first effort in this problem space to use cross-modal, vision-language features for improved pedestrian intent prediction. Besides, while earlier methods largely follow combinations of VGG \cite{Simonyan2015VeryDC} and GRU \cite{cho2014learning1} architectures, we build on contemporary models such as transformers (Swin Transformer V2 \cite{Liu_2022_CVPR}, Sentence-BERT \cite{reimers2019sentence}, and Transformer \cite{vaswani2017attention}) for encoding spatio-temporal features and textual explanation embeddings in this work. 

\section{PIE++: Reason-Enriched Dataset for Pedestrian Intent Estimation}
\label{pie++}
Existing datasets for pedestrian intent prediction provide rich behavior and intent annotations; however, as stated earlier, understanding a pedestrian intent along with its reason enhances reliability in the developed system. Besides, it helps predict the intent itself better, as we show in this work. 
To address this, we propose PIE++, an enrichment of the PIE dataset \cite{rasouli2019pie} with reasons for pedestrian intent. We hypothesize that with the significant advancements of text embeddings in recent years, textual representations of such reasons will support the better prediction of pedestrian intent. We annotate different reasons for both crossing and no-crossing pedestrian intents, as shown in Tables \ref{tab:my_label1} and \ref{tab:my_label2}. 
For pedestrians that: (i) have no crossing intention, there could be reasons such as \textit{``Pedestrians are just doing their work"} or \textit{``Two pedestrians just interacting with each other on the road-side"}; and (ii) have crossing intention, reasons could include  \textit{``Pedestrian acknowledges the ego-car to stop/slow-down with a hand-ack gesture since the pedestrian is intending to cross"}. 
Our list of reasons indicate that it is possible for more than one reason to be relevant for a given pedestrian's intent in a scene. We hence provide \textit{multi-label annotations} for every pedestrian for their crossing vs no-crossing intent. 

\begin{table}[h!]
\centering
\setlength{\tabcolsep}{1pt}

\footnotesize
\begin{tabular}{c | c | c | c | c | c | c }
\hline
\rowcolor[rgb]{0.92,0.92,0.92}
\textbf{Dataset} & \textbf{Year} & \textbf{Len (mins)} & \textbf{Frames} & \textbf{Peds} & \textbf{Intent} & \textbf{Reason} \\
\hline
KITTI \cite{6248074}  & 2012 & 90 & 80,000 & 12,000 & ✗ & ✗ \\
JAAD \cite{rasouli2017they} & 2017 & 46 & 82,000 & 337,000 & ✓ & ✗ \\
BDD100k \cite{Yu_2020_CVPR} & 2018 & 60,000 & 100,000 & 86,047 & ✗ & ✗ \\
PedX \cite{kim2019pedx} & 2019 & - & 10,152 & 14,091 & ✗ & ✗ \\
PIE \cite{rasouli2019pie} & 2019 & 360 & 909,480 & 738,970 & ✓ & ✗ \\
NuScenes \cite{caesar2020nuscenes} & 2020 & 330 & 1,400,000 & 1,400,000 & ✗ & ✗ \\
STIP \cite{liu2020spatiotemporal} & 2020 & 923.48 & 1,108,176 & 3,500,000 & ✓ & ✗ \\
\hline
PIE\textcolor{red}{++} & 2023 & 360 & 909,480 & 738,970 & ✓ & \textcolor{red}{✓} \\
\hline
\end{tabular}
\caption{Comparison of PIE++ with other datasets. 
}
\vspace{-8mm}
\label{tab:comparewithotherdatasets}
\end{table}
We compare PIE++ with other existing datasets in Table \ref{tab:comparewithotherdatasets}. It is evident that considering the reason behind a pedestrian's intention has been unexplored in prior datasets (including PIE).
(Datasets with a $\times$ in the ``Intent'' column have no pedestrian intent annotations.) While STIP and JAAD provide pedestrian intent, We extend the PIE dataset \cite{rasouli2019pie} in this work, since it is reasonably large and has behavioral annotations that were helpful in validating our reason annotations at times. 

\noindent \textbf{Sourcing Reason Annotations in PIE++:}
Before obtaining the multi-label reason annotations in PIE++, 
we provided videos for all the 1842 pedestrians in PIE to 5 annotation professionals with experience in data annotation in the mobility industry. They were provided with all relevant information, including pedestrian intent annotations and pedestrian attributes (e.g. looking, walking, standing) to help decide on the plausible reasons leading to the pedestrian's intent in a given scene. A set of plausible reasons for crossing/no-crossing (C/NC) intent was first developed after viewing the scenes in the dataset. Our reason/explanation categories were decided in a 3-stage process: (i) we first came up with a preliminary set of reason categories based on a user survey; (ii) then thoroughly examined a significant part of the PIE dataset to refine the reason categories; and (iii) added reasons for corner cases (e.g. ``pedestrian neglecting ego-car'', ``pedestrian giving right-of-way to ego-car''). Each of the annotation professionals subsequently went through a given video sequence as many times as required to understand the scene. We asked the annotators to choose as many options as relevant from the set of reason categories for a particular C/NC intent of the pedestrian. We also included an additional option - `other' if a subject considered a different reason. The pedestrian of interest was highlighted with a bounding box in each video frame for consistency of the study. Since we do not have access to ground truth reason data, we analyzed the agreement among annotators to validate our results with this study. First, we computed the intraclass correlation coefficient (ICC), a measure of inter-rater consistency, commonly used to analyze subjective responses from a population of raters in the absence of ground truth \cite{shrout1979intraclass}. Our measured ICC was 0.98, which suggests a very high degree of agreement among our raters for the proposed reason annotations (ICC = 1 for absolute agreement).  
Also notably, the ``\textit{other}" option was never chosen by the annotators in this study, corroborating the sufficiency of the reason categories. 

\begin{table} [t]
\centering
\footnotesize
\begin{tabular}{c }
\hline
\hline
\rowcolor[rgb]{0.92,0.92,0.92}
\textbf{Reasons/Explanations} \\
\hline
\hline
 Waiting to cross with a neighbouring pedestrian \\ 
\hline
 Waiting for a safe passage to cross \\ 
\hline

 Waiting to cross with a group of pedestrians \\ 
\hline
 Waiting for the signal to turn red \\ 
\hline
 Waiting since the ego-vehicle speed is high \\  \hline                                  
 Waiting since the vehicle speed is high \\ 
\hline
 Waiting for vehicles to slow down \\ 
\hline
 Waiting while giving right-of-way to ego-vehicle \\
\hline
 Pedestrian acknowledges ego-vehicle to stop  \\
\hline
 Pedestrian intends to cross since the signal is red \\
\hline
Pedestrian intends to cross since it’s a safe passage \\
\hline
 Pedestrian intends to cross since ego-vehicle speed is slow \\
\hline
Pedestrian intends to cross since vehicle speed is slow \\
\hline 
Neglects the ego-vehicle \\

\hline
\end{tabular}
\vspace{-1mm}
\caption{Reasons used for `\textit{Cross}' pedestrian intent}
\vspace{-6mm}
\label{tab:my_label1}
\end{table}

\begin{table}[t]
\centering
\footnotesize
\begin{tabular}{c }
\hline
\hline
\rowcolor[rgb]{0.92,0.92,0.92}
\textbf{Reasons/Explanations} \\
\hline
\hline
 Two pedestrians just interacting (on road-side) \\ 
\hline
 Group of pedestrians just interacting (on road-side) \\ 
\hline
 Pedestrians doing their work on road-side \\ 
\hline
\hline
\end{tabular}
\vspace{-1mm}
\caption{Reasons used for `\textit{No-Cross}' pedestrian intent}
\vspace{-10mm}
\label{tab:my_label2}
\end{table}

\noindent \textbf{Validating Reason Annotations in PIE++:}
We also validated the goodness of the reason annotations (on parameters of completeness, reliability, trust, correctness, and usefulness) using an independent study in a lab setting. 
This study involved watching video sequences for each pedestrian in the PIE++ dataset. Along with this, we provided the intent and reason annotations to 5 human subjects who were asked to watch each video sequence twice. We then asked the subjects to rate the appropriateness of annotations in PIE++ for the given road scene on a 3-point scale: \textit{Yes}, \textit{Confusing} and \textit{No}.  
The specific questions for each subject included: (1) Are the provided reasons complete for the (No-)cross intent (\textit{completeness})?; (2) Would the reasons be useful if this was provided to you while driving \textit{(usefulness)}?; (3) How correct are the reasons for the pedestrian's perceived intent (\textit{correctness})?; (4) If you were the driver of a car with such predictions, would this improve your trust in the model prediction (\textit{trust})?; (5) How reliable is the provided reason for the prediction (\textit{reliability})? Table \ref{tab:reasonsuserstudy} summarizes the results of our study, and shows strong support for the reason annotations included in PIE++ across the considered factors. 


\begin{table}[t]
\centering
\footnotesize
\setlength{\tabcolsep}{5pt}
\begin{tabular}{|c || c |c | c| c| }
\hline
\rowcolor[rgb]{0.92,0.92,0.92}
\textbf{Factors} & \textbf{Yes} & \textbf{Confusing} & \textbf{No} \\
\hline
Completeness & 100\% & - &  - \\
\hline
Usefulness & 98\% & 1\%  &  1\%   \\
\hline
Correctness & 99\%  & 1\%  & - \\
\hline
Trust & 98\% & 2\%  & - \\
\hline
Reliability & 97\% & 2\%  & 1\%  \\
\hline
\end{tabular}
\vspace{-1mm}
\caption{Results from our user study on the utility of reason annotations in PIE++}
\vspace{-10mm}
\label{tab:reasonsuserstudy}
\end{table}

We note that our work does not change any pre-existing definitions of pedestrian intent defined in \cite{rasouli2019pie}. Also, following \cite{rasouli2019pie}, we consider the intent only up to a critical point (i.e. frames preceding the crossing point). 
PIE++ consists of 4950 multi-label reason annotations for 1410 pedestrians with crossing intent and 657 multi-label reason annotations for 432 pedestrians with no-crossing intent. 
\begin{figure*} 
\begin{center}
\includegraphics[width=0.94\linewidth]{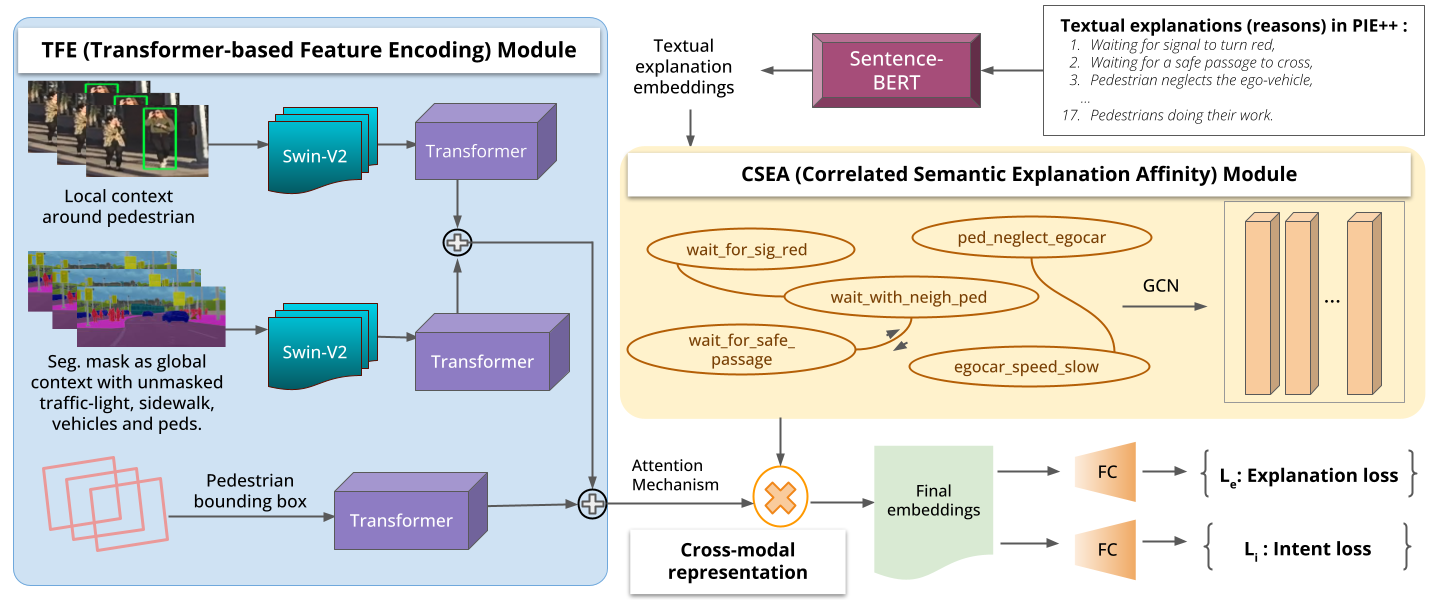}
\end{center}
   \vspace{-5mm}
   \caption{Overall architecture of \textsc{MINDREAD} (cross-Modal representatIon learNing moDel for REAsoning peDestrian-intent)}
   \vspace{-5mm}
\label{fig:arch}
\end{figure*}



\section{\textsc{MINDREAD}: Our Proposed Methodology}
\label{approach}
We propose a multi-task learning architecture with novel components, which we call \textsc{MINDREAD} (cross-Modal representatIon learNing moDel for REAsoning peDestrian-intent), to predict pedestrian intent as well as the reasons behind the intent. Our overall architecture, which seeks to learn a cross-modal representation for the aforementioned objective, is shown in Fig \ref{fig:arch}. 
The cross-modal representation is learned from vision and language modalities, where we use an attention mechanism to fuse vision-based spatio-temporal features and reason text embeddings. 

In this section, we discuss architecture details of our \textsc{MINDREAD} framework. 
Given a video dataset (e.g. a driving dataset containing 2-D RGB traffic video clips), let [$F_1, F_2, ..., F_t$] denote $t$ time steps of past observations (image frames) in a sequence. 
Our goal is to detect the probability of a pedestrian’s crossing intent $I_t \in \{0, 1\}$ ($0=NC, 1=C$) as well as the multi-label textual reasons $ E_t \in \{0, 1\}^{n}$ (Tables \ref{tab:my_label1} and \ref{tab:my_label2}) that support the intent prediction at time $t$; ($n$ represents the total reason classes). 
\textsc{MINDREAD} consists of three modules: Correlated Semantic Explanation Affinity (CSEA), Transformer-based Feature Encoding (TFE) and an attention-based cross-modal representation learning module. The CSEA module receives the textual embeddings of the reasons, and represents them as a learned correlated semantic embedding matrix based on their co-occurrence in the dataset. The TFE module extracts and encodes spatio-temporal features from video frames using a fusion of local features, global features and bounding box information of pedestrians. An attention mechanism brings together the outputs of the TFE and CSEA modules to provide our final cross-modal representation. These cross-modal features are finally fed to two classifier heads that output reason and pedestrian intent predictions (as shown in Fig \ref{fig:arch}). We explain each module of the proposed method in the following subsections.


\subsection{Correlated Semantic Explanation Affinity}
\label{CSEA}
\label{germ}
We model the list of textual reasons as a directed graph where each node is a reason (rather, its text embedding), and the edges denote their co-occurrence in the dataset. We note that since each pedestrian intent instance is annotated with reasons in a multi-label manner, multiple reasons are plausible for a given intent, thus making their co-occurrence a rich piece of input information to the model. While some reasons may or may not appear together, some of them can never appear together. For e.g., reasons like ``\textit{pedestrian neglects ego-vehicle}" can never appear with ``\textit{pedestrians doing their work on the road-side}". On the other hand, reasons like ``\textit{pedestrian waiting since ego-vehicle speed is high}" have a higher chance of appearing with ``\textit{pedestrian waiting for a safe passage}".
Such semantic explanation affinity modeling helps improve reason prediction thereby improving intent prediction. 
We use the widely used Sentence-BERT \cite{reimers2019sentence} to generate textual embeddings for the graph nodes. 
To obtain the co-occurrence of reasons within the dataset, we compute the adjacency matrix using the conditional probability $P(E_j |E_i)$, which denotes the probability of occurrence of reason $E_j$ when reason $E_i$ appears. Given the above explanation node embeddings $X_l \in R^{n \times d}$(where $n$ is the number of reasons and $d$ is the node embedding dimensionality) and adjacency matrix $A \in R^{n\times n}$, we use a Graph Convolution Network (GCN) \cite{welling2016semi} to learn the final reason embeddings. 
Every GCN layer can be written as a non-linear function  $X_{l+1} = f(X_l, A)$, which takes the textual reason representations from the previous layer $(X_l)$ as inputs and outputs new correlated reason representations $X_{l+1}$. For the first layer, the input is an $R^{n \times d}$ matrix, where $d$ is the dimensionality of the reason-level sentence embedding. For the last layer, we choose output of dimension $R^{n \times D}$ where $D$ represents the dimensionality of the output of the TFE module with visual spatio-temporal features.
This gives the final learned correlated semantic explanation embeddings (output of the CSEA module). 

\subsection{Transformer-based Feature Encoding}
\label{tfe}
Following earlier work in intent prediction, the visual spatiotemporal features in \textsc{MINDREAD} are obtained using a combination of local visual context, global visual context and pedestrian bounding box coordinates in the scene (as shown in Fig \ref{fig:arch}). The local context consists of a region around the target pedestrian across the given video sequence (we use a small square patch around the pedestrian in this work). Unlike previous efforts that use pedestrian information alone, we note that global context is useful as it offers visual features that account for socio-environmental interactions. Our global context includes the pixels corresponding to key socio-environmental information in the scene such as traffic-light, vehicles, crosswalks, and other pedestrians (obtained from a segmented scene with masking). 


In order to obtain visual spatio-temporal features from the video frames, we use transformer models \cite{vaswani2017attention} (in contrast to earlier work that use older architectures such as VGG \cite{Simonyan2015VeryDC} and GRU \cite{cho2014learning1} to extract spatial and temporal features). We extract both local and global vision features using two Swin-V2 transformers \cite{Liu_2022_CVPR} pre-trained on the ImageNet-22K \cite{russakovsky2015imagenet} dataset. The final dense layer of the pre-trained model is replaced by a global average pooling layer to extract visual features. Similar to \cite{dosovitskiy2020image1}, we use a standard transformer model for temporal encoding of the features obtained from the Swin-V2 transformers.
The transformer encoder consists of alternating layers of multiheaded self-attention (MSA) and multi-layer perceptron (MLP) blocks. Layernorm is applied before every block, and residual connections after every block following \cite{wang2019learning,baevski2018adaptive}. The MLP contains two layers with a GELU non-linearity. Before feeding the sequence features into the the transformer MSA block, they are combined with positional embedding to capture the sequential properties of input sequence features. 
As in Fig \ref{fig:arch}, both local and global context features are encoded by extracting spatial features from the Swin-V2 transformer followed by temporal features using the Transformer. 
Bounding box features 
are also encoded similarly. We concatenate these three kinds of feature to get the final fused spatiotemporal features. 

\begin{table*} [t]
\centering
\begin{tabular}{|c || c | c| c | c | c| }
\hline
\rowcolor[rgb]{0.92,0.92,0.92}
\textbf{Method} & \textbf{Intent Accuracy} & \textbf{F1-score} &  \textbf{Precision} & \textbf{AUC}  \\
\hline
PIE$_{int}$ (ICCV' 19) \cite{rasouli2019pie} & 79.0 \% & 87.0 \% & 88.0 \% & 79.0 \%  \\
\hline
STIP (ICRA' 20) \cite{liu2020spatiotemporal} & 80.0 \% & 88.0 \% & 91.0 \% & 81.0 \% \\
\hline
CIA (IJCAI' 21) \cite{yao2021coupling} & 82.0 \% & 88.0 \% & 94.0 \% & 83.0 \% \\
\hline
\hline
\textbf{\textsc{MINDREAD} (Ours)} & \textbf{87.6 $\pm$ 0.1\%} & \textbf{95.0 $\pm$ 0.1\%} & \textbf{96.0 $\pm$ 0.1\%} & \textbf{89.0 $\pm$ 0.2\%}  \\

\hline
\end{tabular}
\vspace{-1mm}
\caption{Pedestrian intent prediction on PIE++ (same as PIE for this task)}
\vspace{-6mm}
\label{tab:my_label4}
\end{table*}

\begin{table*} [t]
\centering
\small

\begin{tabular}{|c || c | c | c | c | c |}
\hline
\rowcolor[rgb]{0.92,0.92,0.92}
\textbf{Method} & \textbf{Reason Accuracy} & \textbf{Intent Accuracy} & \textbf{F1-score} & \textbf{Precision} & \textbf{AUC}  \\
\hline
PIE$_{int}$* & 48.0  $\pm$ 0.2\%  & 79.5  $\pm$ 0.1\%   & 87.0 $\pm$ 0.1\% & 88.0 $\pm$ 0.1\% & 79.0 $\pm$ 0.2\% \\
\hline
STIP* & 54.0  $\pm$ 0.1\%  & 80.7  $\pm$ 0.1\%  & 88.0 $\pm$ 0.2\% & 91.0 $\pm$ 0.1\%  &  81.0 $\pm$ 0.1\% \\
\hline
CIA* & 63.0  $\pm$ 0.2\%  & 82.8  $\pm$ 0.1\%  & 88.0 $\pm$ 0.3\% & 94.0 $\pm$ 0.2\%  &  83.0 $\pm$ 0.1\% \\
\hline
\hline
\textbf{\textsc{MINDREAD} (Ours)} & \textbf{72.4  $\pm$ 0.2\% } & \textbf{87.6  $\pm$ 0.1\% } & \textbf{95.0 $\pm$ 0.1\%} & \textbf{96.0 $\pm$ 0.1\%} & \textbf{89.0 $\pm$ 0.2\%}  \\

\hline
\end{tabular}
\vspace{-1mm}
\caption{Pedestrian intent and reason prediction results on PIE++ dataset. Methods with * indicate an adaptation of existing baselines with a reason-based classifier head for a fair comparison.} 
\vspace{-8mm}
\label{tab:my_label6}
\end{table*}

\begin{table}[h!]
\footnotesize
\centering
\begin{tabular}{|c || c | c |}
\hline
\rowcolor[rgb]{0.92,0.92,0.92}
\textbf{Method} & \textbf{Reason Acc} & \textbf{Intent Acc} \\
\hline
PV-LSTM* & 85.3 $\pm$ 0.1 \% & 91.6 $\pm$ 0.1 \% \\
\hline
\hline
\textbf{\textsc{MINDREAD} (Ours)} & \textbf{91.0 $\pm$ 0.1 \%} & \textbf{95.4 $\pm$ 0.1 \%} \\

\hline
\end{tabular}
\vspace{-1mm}
\caption{Pedestrian intent and reason prediction on JAAD. Methods with * indicate adaptation of existing baselines with a reason-based classifier head for fair comparison.}
\vspace{-3mm}
\label{tab:my_label7}
\end{table}

\begin{table} [t]
\centering
\footnotesize
\begin{tabular}{|c || c | c | c |}
\hline
\rowcolor[rgb]{0.92,0.92,0.92}
\textbf{Method} & \textbf{Reason Acc} & \textbf{Intent Acc} & \textbf{F1}  \\
\hline
\textsc{MINDREAD}\_w/o\_CM & 64.0  $\pm$ 0.1\%  & 83.4  $\pm$ 0.1\%  & 90.0\%\\
\hline
\textbf{\textsc{MINDREAD} (Ours)} & \textbf{72.4  $\pm$ 0.2\% } & \textbf{87.6  $\pm$ 0.1\% } & \textbf{95.0\%}  \\
\hline
\end{tabular}
\vspace{-3mm}
\caption{\textsc{MINDREAD} with and without cross-modality}
\label{tab:ablation3table}
\vspace{-7mm}
\end{table}

\subsection{Attention-based Cross-modal Representations} 
The final module of our framework, an attention-based mechanism to fuse the outputs of the TFE and CSEA modules, helps leverage the reason relationships to better predict pedestrian intent. This is a key element of our framework. The TFE output features are fed to an attention block for selectively focusing on the appropriate feature information for our task. The temporal sequence of features (output of transformer-based encoding from the TFE module) are represented as hidden states $h = {h_1,h_2,...,h_e}$, where $e$ represents the end hidden state. Following \cite{luong2015effective} , the attention weight is computed 
as: $\alpha = \frac {exp(score(h_e, \Tilde{h_s}))} { \Sigma_{s'} exp(score(h_e, \Tilde{h_{s'}}))} $, where $score(h_e,\Tilde{h_s}) = h^T_eW_s\Tilde{h_s}$ and $W_s$ is a learned weight matrix. Such an attention mechanism provides information on how the current video frame (i.e. the end hidden state $h_e$) should leverage the immediately preceding frames given in $\Tilde{h_s}$. 
The final output of the attention module is obtained as $F = \tanh(W_c[h_c;h_e])$, where $W_c$ is a weight matrix, $h_c$ is the sum of all attention-weighted hidden states given as $h_c = \Sigma_{s'} \alpha \Tilde{h_{s'}}$ and $F \in R^{B\times 3076}$, where $B$ represent batch size. 
Finally, our cross-modal representation is given by, $C = F \cdot X^T$, where $F$ represents final attention-based feature representations and $X$ represents the learned correlated semantic explanation embeddings.
These final cross-modal encodings are fed to two classifiers giving textual explanation $E_t$ and intent $I_t$ prediction outputs. We use standard Binary Cross-Entropy (BCE) losses, $L_e$ for explanation and $L_i$ for intent, to learn the final parameters of our model. 

\section{Experiments and Results}
We evaluate the effectiveness of our method for intent and reason prediction on PIE++. To study the generalizability of this approach, we also annotate the JAAD \cite{rasouli2017they} dataset with reason explanations, although this dataset has very short sequences with limited variety in user behavior.
We first present our results on the pedestrian intent prediction task, where we compare our method's performance with state-of-the-art methods, followed by results for the novel problem of reasoning the pedestrian intent. Following earlier methods in this space, we measure performance using intent accuracy, explanation accuracy, and F1 score. Qualitative results on video sequences are provided in the supplementary video. 


We compare our method with well-known methods for pedestrian intent prediction: PIE$_{int}$ \cite{rasouli2019pie}, STIP \cite{liu2020spatiotemporal} and CIA \cite{yao2021coupling} on PIE++, and PV-LSTM \cite{bouhsain2020pedestrian} on the JAAD dataset. 
Table \ref{tab:my_label4} shows the intent prediction results on PIE++ dataset. \textsc{MINDREAD} outperforms the state-of-the-art for intent prediction, with an improvement in accuracy and F1-score by $5.6\%$ and $7\%$ respectively. 
Table \ref{tab:my_label5} shows results on JAAD. Once again, our method outperforms all earlier works with an improvement of $4.4\%$ over the state-of-the-art on intent prediction accuracy. 

\subsection{Results on Reasoning Pedestrian Intent}
Since there is no existing method for simultaneous intent and reason prediction, we adapt existing methods by adding an explicit reason head and retraining them on PIE++ for fair comparison. We denote the corresponding variants of these methods as PIE$_{int}$*, STIP*, CIA* for the PIE++ dataset, and PV-LSTM* for the JAAD dataset (enriched with our reason annotations). 
\begin{wraptable}[5]{r}{4.8cm}
\vspace{-2mm}
\footnotesize
\begin{tabular}{|c || c | }
\hline
\rowcolor[rgb]{0.92,0.92,0.92}
\textbf{Method} & \textbf{Intent Acc} \\
\hline
PV-LSTM \cite{bouhsain2020pedestrian} & 91.0 \% \\
\hline
\hline
\textbf{\textsc{MINDREAD} (Ours)} & \textbf{95.4 \%} \\

\hline
\end{tabular}
\vspace{-4mm}
\caption{Intent prediction on JAAD}
\vspace{-2mm}
\label{tab:my_label5}
\end{wraptable}
Table \ref{tab:my_label6} shows pedestrian intent and reason prediction results on PIE++. \textsc{MINDREAD} outperforms all the baseline methods, including state-of-the-art method CIA* with significant improvement of $9.4\%$ in reason prediction accuracy, $4.8\%$ in intent prediction accuracy and $7\%$ in intent F1-score. We hypothesize that this improvement in performance is due to our cross-modal representation learning as is also evident from our ablation studies (Sec \ref{ablation} Table \ref{tab:ablation3table}). Table \ref{tab:my_label7} shows results of similar experiments on the JAAD dataset. \textsc{MINDREAD} improves over the state-of-the-art method by $5.7\%$ and $3.8\%$ improvement in reason and intent accuracy. For JAAD, we use the same train-test split as \cite{8317941} and F1-score, precision, and AUC are 97\%, 95\% and 91\%. 

\begin{table}[t]
\centering
\footnotesize

\begin{tabular}{|c || c | c| c | c | c| }
\hline
\rowcolor[rgb]{0.92,0.92,0.92}
\textbf{Textual Emb.} & \textbf{Reason Acc} & \textbf{Intent Acc} & \textbf{F1}  \\
\hline
GloVe & 68.2  $\pm$ 0.1\%  & 86.4  $\pm$ 0.1\%  & 93.0 \%  \\
\hline
Sentence-BERT & \textbf{72.4  $\pm$ 0.2\% } & \textbf{87.6  $\pm$ 0.1\% } & \textbf{95.0 \%}  \\

\hline
\end{tabular}
\vspace{-1mm}
\caption{Study comparing word-level (GloVe) vs sentence-level  (Sentence-BERT) models for reason embeddings}
\vspace{-5mm}
\label{tab:my_label9}
\end{table}


\vspace{-2mm}
\begin{table}[t] 
\centering
\footnotesize

\begin{tabular}{|c || c | c| c| }
\hline
\rowcolor[rgb]{0.92,0.92,0.92}
\textbf{Backbone} & \textbf{Exp. Acc} & \textbf{Intent Acc} & \textbf{F1} \\
\hline
VGG + GRU & 69.2  $\pm$ 0.1\%  & 87.0  $\pm$ 0.1\%  & 94.0\% \\
\hline
Swin-V2 + Transformer & \textbf{72.4  $\pm$ 0.2\% } & \textbf{87.6  $\pm$ 0.1\% }  & \textbf{95.0\%} \\
\hline
\end{tabular}
\vspace{-3mm}
\caption{Ablation using VGG + GRU versus novel Swin-V2 + transformer backbone for spatio-temporal visual feature modeling.}
\vspace{-7mm}
\label{tab:my_label8}
\end{table}


\vspace{-2mm}
\begin{table}[t] 
\centering
\footnotesize
\begin{tabular}{|c || c | c| }
\hline
\rowcolor[rgb]{0.92,0.92,0.92}
\textbf{Method} & \textbf{Runtime} & \textbf{Intent Acc} \\
\hline
\textsc{PV-LSTM \cite{bouhsain2020pedestrian}} & 4.7 ms & 91.0 \% \\
\hline
\textbf{\textsc{MINDREAD}} & \textbf{4.3 ms} & \textbf{95.4 \%}  \\
\hline
\end{tabular}
\vspace{-1mm}
\caption{Run time analysis on JAAD dataset}
\vspace{-6mm}
\label{tab:runtime_analysis}
\end{table}


\vspace{-2mm}
\begin{table}[t] 
\centering
\footnotesize
\begin{tabular}{|c | c | c| c | }
\hline
\rowcolor[rgb]{0.92,0.92,0.92}
\textbf{$\gamma_{R}$} & \textbf{$\gamma_I$} & \textbf{Reason Acc} & \textbf{Intent Acc} \\
\hline
0.5 & 1.0 & 57.4 \% & 81.2 \% \\
\hline
1.0 & 0.5 & 63.6 \% & 74.1 \% \\
\hline
\textbf{1.0} & \textbf{1.0} & \textbf{72.4 \%} & \textbf{87.6 \%} \\
\hline
\end{tabular}
\vspace{-1mm}
\caption{Results using different weights; $\gamma_r$ = weight for reason loss; $\gamma_i$ = weight for intent loss}
\vspace{-10mm}
\label{tab:mtlweights}
\end{table}

\vspace{7mm}
\section{Analysis and Discussion}
\label{ablation}
\textbf{Word vs. Sentence-level Embeddings.}
We studied the relevance of different text embedding methods, as shown in the results in Table \ref{tab:my_label9}. Specifically, we compare word vs sentence embedding models. For word-level embedding, we use a 300-dim Glove \cite{pennington2014glove} model trained on the Wikipedia dataset. We consider reason-specific words and average the embeddings. For sentence-level embeddings, we use Sentence-BERT \cite{reimers2019sentence} (as in our method). The results show that sentence-level embeddings outperform word-level embeddings. This is expected since the context of the words in the reason matter, which is considered in sentence-level embeddings. 

\textbf{Effect of Different Spatio-temporal Backbones.}
Table \ref{tab:my_label8} shows the results of our study using different spatio-temporal visual feature extraction backbones in \textsc{MINDREAD}. As in Sec \ref{approach}, we use contemporary transformer models instead of older models used in earlier work. Specifically we use Swin-V2-L + Transformer backbone instead of the VGG + GRU backbone. In order to study this choice, we used the VGG +GRU backbone in \textsc{MINDREAD} and obtained the results. Table \ref{tab:my_label8} shows that transformer-based models outperform VGG + GRU models by $3.2\%$ in reason prediction accuracy and $0.6\%$ in intent prediction accuracy. 

\textbf{\textsc{MINDREAD} without Cross-modality.} 
In order to understand the usefulness of the cross-modal representation learning framework proposed in \textsc{MINDREAD}, we conducted experiments to study whether visual features alone can address the new task with an added reason classification head (in addition to the intent prediction head). 
As shown in Table \ref{tab:ablation3table}, we can see performance drop by 4.2\% and 5\% in intent accuracy and F1 using \textsc{MINDREAD} without cross-modality. We observe that the use of reason embeddings and cross-modal representation learning plays a significant role for reason prediction accuracy in particular, and also helps improve intent prediction (as we hypothesized).
\textbf{Loss, Metrics and Runtime Analysis.} 
 Our main results used a weight of 1 for each loss term. Our initial studies, reported in Tab \ref{tab:mtlweights}, supported this choice. Weighting one loss term more than the other did not increase performance. 
Tab \ref{tab:runtime_analysis} shows the runtimes, compared with those reported on JAAD. We note that our model is faster than SOTA PV-LSTM \cite{bouhsain2020pedestrian} while achieving higher accuracy. Runtime for our model on PIE++ was 4.1 ms, of the same range as JAAD. It's due to the significance of \textit{learned} reason embeddings as well as spatiotemporal features that speed up inference in real time. 

\section{Conclusions and Future Work}
We introduced a novel problem of reasoning pedestrian intent. To address this problem, we made two significant contributions: (1) We proposed a reason-enriched PIE++ dataset that contains explanations/textual reasons for pedestrian intent prediction; and (2) We proposed a multi-task formulation called \textsc{MINDREAD} based on cross-modal representation learning that obtains strong results on this new problem setting. 
Our comprehensive experiments show we achieve significantly more accurate and reasonable predictions than prior works, and that such explanations improve trust in users of such systems. Exploring uncertainty estimates for reasons and collision avoidance are interesting future directions of our work. 


\bibliographystyle{IEEEtran}
\bibliography{IEEEabrv,IEEEfull}

\end{document}